\pdfoutput=1

\documentclass[11pt]{article}

\usepackage{acl}

\usepackage{times}
\usepackage{latexsym}
\usepackage{subfigure}
\usepackage{algorithm}
\usepackage{algorithmic}
\usepackage{makecell}
\usepackage{color}
\usepackage{xcolor}

\usepackage{graphicx}
\usepackage{booktabs}
\usepackage{amssymb}
\usepackage{amsmath}
\usepackage{appendix}
\usepackage{enumitem}
\usepackage{multicol}
\usepackage{multirow}
\usepackage[normalem]{ulem}
\useunder{\uline}{\ul}{}
\usepackage{colortbl}
\usepackage{tabularx}
\usepackage{booktabs}
\usepackage{tabularx}
\usepackage[T1]{fontenc}
\usepackage{color}

\usepackage[utf8]{inputenc}
\usepackage{pdfpages}

\usepackage{microtype}

\title{LEVEN: A Large-Scale Chinese Legal Event Detection Dataset}

\newcommand{\dataset}{LEVEN}

\author{Feng Yao\textsuperscript{\rm 1}{$^*$},
Chaojun Xiao\textsuperscript{\rm 2,3}\thanks{~~Equal contribution. Listing order is random.}~, 
Xiaozhi Wang\textsuperscript{\rm 2,3},
Zhiyuan Liu\textsuperscript{\rm 2,3,4,5}\thanks{~~Corresponding authors.}~,\\
\textbf{
Lei Hou\textsuperscript{\rm 2,3},
Cunchao Tu\textsuperscript{\rm 6},
Juanzi Li\textsuperscript{\rm 2,3},
Yun Liu\textsuperscript{\rm 1},
Weixing Shen\textsuperscript{\rm 1}{$^\dagger$},
Maosong Sun\textsuperscript{\rm 2,3,4,5}} \\

\textsuperscript{\rm 1}School of Law, Institute for AI and Law, Tsinghua University, Beijing, China \\
\textsuperscript{\rm 2}Dept. of Comp. Sci. \& Tech., Institute for AI, Tsinghua University, Beijing, China \\
\textsuperscript{\rm 3}Beijing National Research Center for Information Science and Technology, China \\
\textsuperscript{\rm 4}International Innovation Center of Tsinghua University, Shanghai, China \\
\textsuperscript{\rm 5}Beijing Academy of Artificial Intelligence, Beijing, China \\
\textsuperscript{\rm 6}Beijing Powerlaw Intelligent Technology Co., Ltd., China \\
\texttt{\{yaof20,xiaocj20\}@mails.tsinghua.edu.cn} \\
\texttt{\{liuzy,wxshen\}@tsinghua.edu.cn}}

\begin{document}
\maketitle
\begin{abstract}
Recognizing facts is the most fundamental step in making judgments, hence detecting events in the legal documents is important to legal case analysis tasks. However, existing Legal Event Detection (LED) datasets only concern incomprehensive event types and have limited annotated data, which restricts the development of LED methods and their downstream applications. To alleviate these issues, we present \dataset{}, a large-scale Chinese LEgal eVENt detection dataset, with $8,116$ legal documents and $150,977$ human-annotated event mentions in $108$ event types. Not only charge-related events, \dataset{} also covers general events, which are critical for legal case understanding but neglected in existing LED datasets. To our knowledge, \dataset{} is the largest LED dataset and has dozens of times the data scale of others, which shall significantly promote the training and evaluation of LED methods. The results of extensive experiments indicate that LED is challenging and needs further effort. Moreover, we simply utilize legal events as side information to promote downstream applications. The method achieves improvements of average 2.2 points precision in low-resource judgment prediction, and 1.5 points mean average precision in unsupervised case retrieval, which suggests the fundamentality of LED. The source code and dataset can be obtained from~\url{https://github.com/thunlp/LEVEN}.

\end{abstract}

\section{Introduction}
\begin{figure}[t]
    \centering
    \includegraphics[width=\columnwidth]{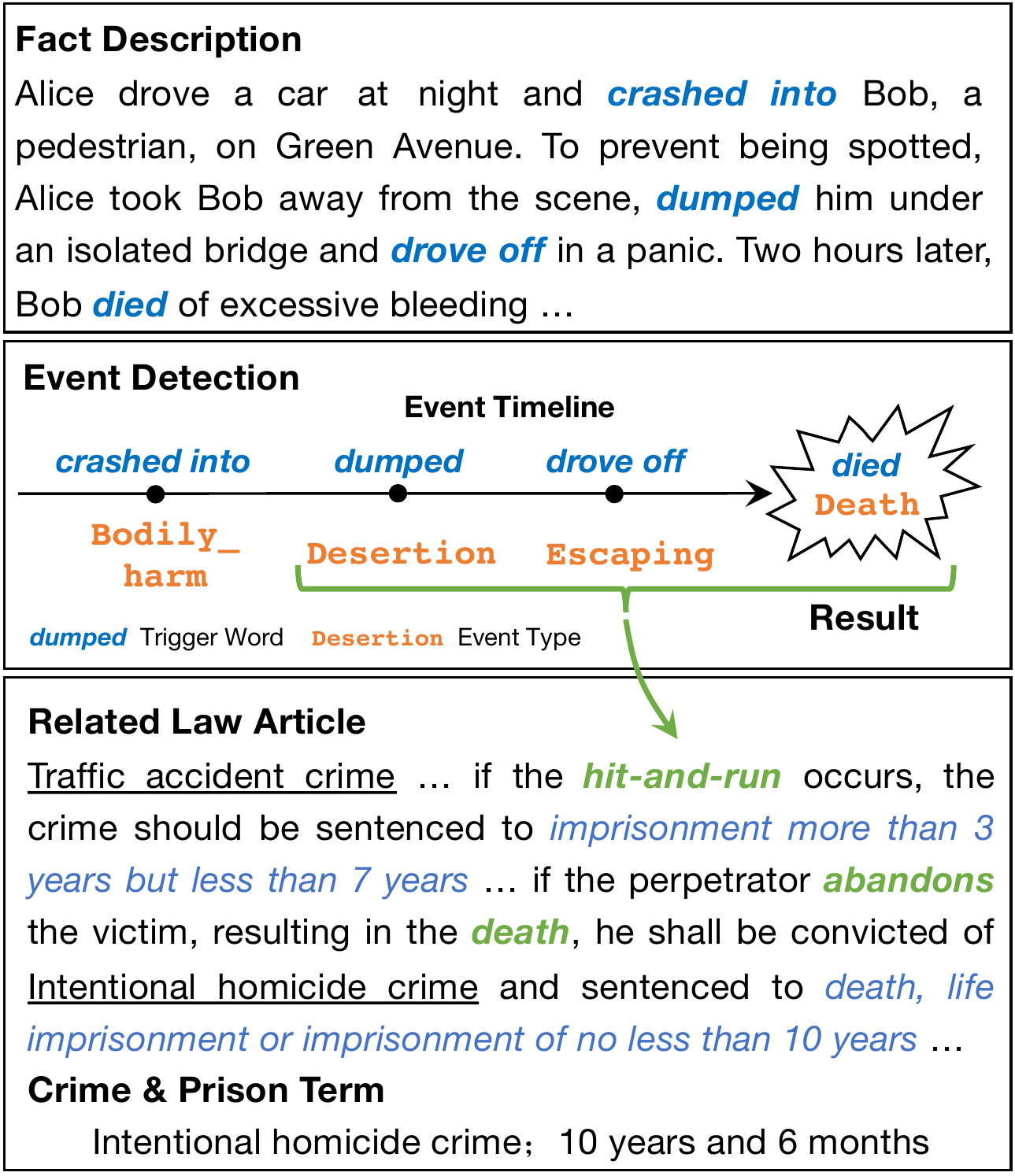}
    \caption{An example legal document describing the fact with the annotated event triggers, the corresponding event types, the related law article, and penalties.}
    \label{fig:example}
\end{figure}

Finding out the occurred events and causal relations between them is fundamental to analyzing legal cases and making judgments. Legal event detection (LED) aims to automatically extract the event triggers from legal cases and then classify their corresponding event types, which will naturally benefit many legal artificial intelligence applications, such as Legal Judgment Prediction (LJP) and Similar Case Retrieval (SCR)~\cite{zhong2020does}. For instance, Figure~\ref{fig:example} shows a case with the trigger words highlighted in the plain text and the corresponding event types. Based on the detected events, we can observe that Alice causes a traffic accident, and the subsequent \texttt{Desertion} and \texttt{Escaping} events jointly result in the \texttt{Death} event, which changes Alice's charge from traffic accident crime to intentional homicide crime and increases the expected penalties.

Inspired by the previous success for general-domain event detection~\cite{ji2008refining,li2013joint,chen2015event,nguyen2016joint,feng2018language,yan2019event,wang2020maven}, some works attempt to build LED systems with hand-crafted features~\cite{lagos2010event,bertoldi2014cognitive}, or neural networks~\cite{li2019apply,li2020event,shen2020hierarchical}. However, two major challenges of existing LED resources seriously restrict the development of LED methods:

(1) \textbf{Limited Data.} Existing LED datasets~\citep{shen2020hierarchical,li2020event} only contain thousands of event mention annotations, which can not provide sufficient training signals and reliable evaluation results. To promote the progress of legal information extraction and legal document analysis, it is an urgent need to develop a large-scale and high-quality dataset for the LED task. (2) \textbf{Incomprehensive Event Schema.} Existing LED works merely concern a dozen of charge-oriented event types, which are either the judicial event types defined in general-domain datasets~\citep{maxwell2009evaluation} or some newly-defined charge-oriented event types to meet specific downstream requirements~\citep{li2019apply,li2020event,shen2020hierarchical}. Their event schemata only cover a narrow scope of charges. Besides, existing datasets focus on charge-oriented events and ignore the general events in the cases, such as \texttt{Desertion} and \texttt{Escaping} in Figure~\ref{fig:example}, which are also critical for analyzing legal cases.

To alleviate the above issues and provide a solid foundation for LED, we present \dataset{}, a large-scale Chinese legal event detection dataset, based on the cases published by the Chinese government\footnote{\url{https://wenshu.court.gov.cn/}}. We highlight \dataset{} with the following merits:

(1) \textbf{Large scale}. \dataset{} contains $8,116$ legal documents covering $118$ criminal charges and has $150,977$ human-annotated event mentions, which is dozens of times larger than previous LED datasets.
To the best of our knowledge, \dataset{} is also the largest Chinese event detection dataset. Based on the scale, we believe \dataset{} can well train and reliably benchmark data-driven LED methods, which shall promote this field.
(2) \textbf{High coverage}. \dataset{} contains $108$ event types in total, including $64$ charge-oriented events and $44$ general events. The \dataset{} event schema has a sophisticated hierarchical structure, which is shown in appendix~\ref{app:event_schema}. To build the schema, we conduct a two-stage event schema construction process. We first summarize the critical charge-oriented event types based on law articles and then simplify and supplement the event schema based on the events in sample cases. The two-stage process ensures the high coverage of \dataset{} schema.

To explore the challenges of \dataset{}, we implement some state-of-the-art models and evaluate them on our dataset. The results show that though existing models can achieve better performance on legal documents than in the general domain, it still needs future efforts to reach a practical level.

Moreover, we demonstrate the fundamentality of LED for downstream Legal AI applications. Specifically, we train an LED model on \dataset{} and use it to detect events for unlabeled legal documents. We then use the auto-detected events as side information to handle LJP and SCR. Experiments show that the performance of these two tasks can be substantially improved in this simple way, indicating that LED can provide helpful fine-grained information and thus serve as a fundamental process in Legal AI. Hence we advocate more research attention to LED. 

\section{Related Work}
\subsection{Event Detection}
Event detection (ED) is an important information extraction task and many efforts have been devoted to~\cite{ji2008refining,li2013joint,chen2015event,nguyen2016joint,liu2017exploiting,zhao2018document,yan2019event,wang2021cleve}. The majority of existing ED datasets are developed for the general domain~\citep{ace2005,song2015light,wang2020maven} and mostly for English. Besides, some datasets are also developed for specific domains~\citep{thompson2009construction,kim2008text,ritter2012open,yang2018dcfee,zheng2019doc2edag} and Chinese~\citep{li2020duee}. Considering the rapid growth of Chinese legal artificial intelligence~\cite{zhong2020does}, we believe constructing Chinese LED datasets is helpful and necessary. In the context of LED, some works define specific legal event types to analyze for legal documents~\cite{maxwell2009evaluation,lagos2010event,li2019apply,shen2020hierarchical,li2020event}, but these constructed datasets are typically small-scale and cannot well train and evaluate practical LED systems. Hence we construct \dataset{}, which is the largest LED dataset and also the largest Chinese event detection dataset to our knowledge.

\subsection{Legal Artificial Intelligence}
Thanks to the rapid progress of natural language processing and the openness of legal documents, legal artificial intelligence (LegalAI) has drawn increasing attention from both AI researchers and legal professionals in recent years~\cite{bommarito2021lexnlp,ye2018interpretable,chalkidis2021lexglue,zhong2020does,tsarapatsanis2021ethical,wang2021equality}. LegalAI can not only provide handy references for people who are not familiar with legal knowledge, but also reduce the redundant paperwork for legal practitioners. Many efforts have been devoted to a variety of LegalAI tasks, including legal judgment prediction~\cite{zhong2018legal,chalkidis2019neural,yang2019legal}, legal question answering~\cite{ravichander2019question,zhong2020jec,kien2020answering}, contract review~\cite{hendrycks2021cuad,zhang2021learning,koreeda2021contractnli}, legal case retrieval~\cite{ma2021lecard,shao2021investigating}, and legal pre-trained models~\cite{chalkidis2020legal,xiao2021lawformer}. Most existing works focus on the application in LegalAI while ignoring the basic key event information in the legal documents. 
Some works attempt to extract events from the legal documents~\cite{li2019apply,shen2020hierarchical,li2020event}. But these works are limited to the event coverage and the number of annotation instances.
We argue that our proposed large-scale and comprehensive dataset, \dataset{}, can promote the development of legal event detection and thus benefit downstream legal artificial intelligence tasks.

\section{Data Collection}
Our ultimate goal is to construct a large-scale legal event detection dataset with a high-coverage event type schema and sufficient event instances, which is scarce in existing LED datasets. Therefore, we need to redefine a new event schema, select the trigger candidates, and annotate the corresponding event types. As criminal cases usually involve principal rights and complex facts, we focus on criminal legal events in this paper.
In the following sections, we first introduce the construction of event schema and then describe the process of annotation of candidates and related event types.

\subsection{Event Schema Construction}
\label{sec:schema}
To construct an event schema with high coverage, we need to consider events for both judicial behaviors and general behaviors. Therefore, we follow a two-stage process to define our new event schema: 1) We first collect charge-oriented events based on the law articles and legal textbooks. 2) We then collect general events from the sampled case documents. The two-stage process enables \dataset{} to cover essential events recorded in legal documents.

Inspired by previous works~\cite{li2020event,shen2020hierarchical}, in the first stage, we use law articles and a classical legal textbook, \textit{Specific Theory of Criminal Law}, as our references to summarize the charge-oriented events. \textit{Criminal Law} provides the definition of each criminal charge and a hierarchical structure for these charges. We first collect $459$ criminal charges, which are then divided into $61$ types based on the targets and measures of criminal behaviors. Considering that some criminal charges are too abstract to be specific event types (e.g., \texttt{dereliction}), we manually filter out them. Besides, as there are some similar charges involving the same event types (e.g., \texttt{intentional\_homicide} and \texttt{involuntary\_homicide}), we merge them. After the first stage, we obtain $198$ event types highly correlating to the criminal charges.

As the charge-oriented event schema is constructed from legal professional references, there are two main issues: 1) The charge-oriented event schema mainly focuses on illegal behaviors, while ignoring important general behaviors. 2) There are some event types that infrequently or never occur in real-world cases. To address these issues, we further modify the event schema based on the summarization of real-world cases. Specifically, we sample $20$ case documents for each criminal charge, which can ensure good coverage. And then we invite a legal expert to manually extract and summarize the event mentions occurring in sampled cases. Based on the extracted events, we further filter out the abstract event types and merge some detailed event types in the schema. We finally get $108$ event types to annotate, with both charge-oriented events and general events. 

\begin{table*}[ht]
    \centering
    \small
    \begin{tabular}{l|rrrrrrr}
    \toprule
    Dataset  & \#Documents & \#Tokens & \#Sentences & \#Event Types & \#Event Mentions & Language & Domain \\ \midrule
    MAVEN       & 4,480 & 1,276k & 49,873 & 168 & 118,732 & English & General \\
    ACE2005-zh  & 633 & 185k & 7,955 & 33 & 4,090 & Chinese & General \\
    DuEE & 11,224 & 530k & 16,900 & 65 & 19,640 & Chinese & General \\
    DivorceEE* & 3,100 & -- & -- & 13 & -- & Chinese & Legal\\
    CLEE*    & 3,000 & -- & 6,538 & 5 & 6,538 & Chinese & Legal \\
    DyHiLED* & -- & -- & -- & 11 & 2,380 & Chinese & Legal \\
    \midrule
    \dataset{} & 8,116 & 2,241k & 63,616 & 108 & 150,977 & Chinese & Legal \\ \bottomrule
    \end{tabular}
    \caption{The statistics of widely-adopted event detection datasets. For Chinese datasets, we adopt JIEBA toolkit to perform tokenization. Datasets denoted with * are not publicly available, and -- means the value is not accessible.}
    \label{tab:stat}
\end{table*}

According to the criminal theory, the key elements of the crime include the act, the harmful results, and the causal relation between them. Therefore, we organize the event types in a hierarchical structure, with three categories representing behavior and a category representing results.
During the annotation process, the annotators are required to label the most fine-grained types. Please refer to Appendix \ref{app:event_schema} for details of the event schema.

\subsection{Document Selection}
To support the manual annotation, we adopt cases collected from the government website as our data source. Following \citet{xiao2018cail2018}, we only keep the criminal judgment documents for annotation. 

We first extract the related charges with regular expression from the documents and divide each document into several parts based on the content, where only the fact description is maintained. Moreover, to ensure the dataset quality, we filter out the documents with less than $50$ characters and more than $2,500$ characters in fact description.
Notably, though we get $198$ charges in the first stage of event schema construction, there are some charges where no cases are published due to the privacy and secrecy involved. Therefore, we get case documents with only $107$ charges.
We randomly sample $200$ documents for charges with high frequency and maintain all cases for charges with low frequency. Finally, we select $8,288$ documents for annotation. After discarding the low-quality documents labeled by annotators, we finally retain $8,166$ documents.

\subsection{Candidate Selection} \label{sec:cand-select}
The annotation of LED dataset requires the annotators to find the triggers from the documents and label the corresponding event types within $108$ options. Following \citet{wang2020maven}, we adopt heuristic methods to automatically select the trigger candidates and narrow down the event type options for each trigger candidate.

\textbf{Candidate trigger selection.} Inspired by~\citet{chen2017automatically}, which utilizes the lexical unit in FrameNet~\cite{baker1998berkeley} to select trigger words, we require a legal expert to collect semantic-related words for each event type in our schema. And we obtain a semantic vocabulary consisting of $1,013$ words with their corresponding event types. Then we apply tokenization and POS tagging with JIEBA toolkit\footnote{\url{https://github.com/fxsjy/jieba}}, and all the content words, including nouns and verbs, are selected as trigger candidates. Besides, the words in the collected vocabulary are also selected as trigger candidates.

\textbf{Candidate event type selection.} Further, we recommend $30$ event types for each trigger candidate, which can provide references for annotators. 
We first calculate the cosine similarity between the representations of trigger candidates and event types. And then we rank the event types by the calculated similarity and retrieve the top 30 ones as the recommended candidates. Here, we utilize the representations calculated by SBERT~\cite{reimers2019sentence}, which can generate semantic meaningful embeddings.

The automatic candidate selection mechanism aims to provide a good reference for the annotators. Notably, considering that not all triggers and event types can be automatically selected, we also require the annotators to label the words and event types that are not in the recommended list.
The final annotation results show that $95.6\%$ trigger words and $92.8\%$ event types are recommended, and the rest are supplemented by annotators manually.
The results demonstrate that the automatic candidate selection is helpful to improve the annotation efficiency, and the annotators can also label the trigger words and event types that are not recommended.

\begin{table*}[h]
    \centering
    \small
    \resizebox{\textwidth}{!}{
    \begin{tabular}{l|l|rrr|l}
    \toprule
        \makecell{Top-level Event Type} & Category & \#Type & \#Mention & Percentage & \makecell{Sub-type Examples} \\ \midrule
        \texttt{General\_behaviors}     & Behavior & 40 & 68,616 & $45.4\%$ & 
            \texttt{Selling}, \texttt{Employing}, \texttt{Manufacturing} \\
        \texttt{Prohibited\_acts}     & Behavior & 40 & 43,021 & $28.5\%$ & 
            \texttt{Killing}, \texttt{Blackmail}, \texttt{Theft}, \texttt{Destroying} \\
        \texttt{Judicature\_related}    & Behavior & 13 & 29,709 & $19.7\%$ & 
            \texttt{Arrest}, \texttt{Surrendering} \\
        \texttt{Consequences}            & Result & 7 & 6,832 & $4.5\%$ & 
            \texttt{Death}, \texttt{Injury}, \texttt{Being\_trapped} \\
        \texttt{Accident}               & Result & 4 & 2,742 & $1.8\%$ &
            \texttt{Traffic\_accident}, \texttt{Fire\_accident} \\ 
        \texttt{Natural\_disaster}      & Majeure & 4 & 57 & $0.03\%$ & 
            \texttt{Drought},  \texttt{Flood\_and\_waterlogging} \\ 
        \bottomrule
    \end{tabular}
    }
    \caption{Data distribution over the top-level event types and the corresponding categories and samples.}
    \label{tab:data-distribution}
\end{table*}

\subsection{Human Annotation}
The final process is to annotate the triggers from documents manually. We write a 59-page annotation guideline in Chinese to help the annotators better understand the annotation task. We also embed the guideline in the annotation platform so that the annotators can easily refer to it. A simplified version in English is provided in Appendix \ref{app:annote_guide}.

Following previous works~\cite{ace2005,wang2020maven}, we adopt a two-stage annotation process. In the first stage, we invite crowd-source annotators to choose the correct answers from case documents when given the candidate triggers and corresponding event types. Each document is annotated independently twice. The annotators first went through several hours of training for labeling, so as to ensure the annotation quality. Besides, for each labeled document, we discard the annotation results and require another two annotators to annotate it, if the inter-annotator agreement of the document is lower than $0.2$. In the second stage, we invite experienced annotators to choose final event types given the results of the first-stage annotation. Only the results labeled differently in the first stage are required to be labeled again in the second stage.

We measure the data quality via inter-annotator agreements between two annotators, i.e., Cohen's Kappa~\cite{cohen1960coefficient}. The Kappa coefficient for the first stage is $0.609$. To evaluate the data quality in the second stage, we randomly sample $5\%$ documents to be labeled twice independently. The Kappa coefficient for the second stage is $0.875$. The satisfactory Kappa coefficient demonstrates that \dataset{} is a high-quality manually annotated LED dataset, and we hope the dataset can accelerate the development of LED and legal case analysis.

\section{Data Analysis}
In this section, we conduct analysis from various aspects to provide a deep understanding of \dataset{}.

\subsection{Data Size}
The detailed statistics of \dataset{} and some widely-used event detection datasets are shown in Table~\ref{tab:stat}. We compare \dataset{} with two types of datasets: 
(1) \textbf{General-domain ED datasets.} ACE2005~\citep{ace2005} is the most popular multi-lingual event extraction dataset and here we compare with its Chinese subset (denoted as ACE2005-zh). MAVEN~\citep{wang2020maven} is the largest general-domain event detection dataset, with $168$ event types and hundreds of thousands of event instances. DuEE~\citep{li2020duee} is the largest Chinese ED dataset, which is collected from Chinese news articles. 
(2) \textbf{LED datasets.} DivorceEE~\cite{li2019apply} focuses on event extraction in divorce cases. CLEE~\cite{li2020event} is for larceny cases. DyHiLED~\cite{shen2020hierarchical} is a LED dataset with a hierarchical event schema. 
From the comparisons, we can observe that \dataset{} is the largest LED dataset with dozens of the scale of previous LED datasets and is also the largest Chinese event detection dataset. \dataset{}'s scale is even comparable to the previous largest general-domain event detection dataset MAVEN. Moreover, \dataset{} contains the most event types among the Chinese event detection datasets. These suggest that \dataset{} may help LED, Chinese ED, and general ED at the same time.

\begin{table}[h]
    \centering
    \small
    \scalebox{0.95}{
    \begin{tabular}{l|rrrr}
    \toprule
          & \#Doc. & \#Sentences & \#Event & \#Negative. \\ \midrule
    Training & 5,301 & 41,238 & 98,410 & 297,252 \\
    Validation & 1,230 & 9,788 & 22,885 & 69,645 \\
    Test  & 1,585 & 12,590 & 29,682 & 90,512 \\ \bottomrule
    \end{tabular}
    }
    \caption{The detailed statistics of subsets of \dataset{}.}
    \label{tab:train-valid-test}
\end{table}

\subsection{Data Distribution}
\textbf{Event Types}. As mentioned before, our event schema contains three event categories representing behavior, two event categories representing results, and one event category representing force majeure. The instance distribution over these top-level event types is shown in Table~\ref{tab:data-distribution}. There are $45.4\%$ events belonging to general behavior, which is ignored in the previous LED dataset. It demonstrates that modeling the general events in LED is necessary. Besides, \dataset{} meets long-tail distribution, which raises a challenge for future research.

\begin{table*}[t]
    \centering
    \small
    \begin{tabular}{l|ccc|ccc}
    \toprule
    \multirow{2}{*}{Model} & \multicolumn{3}{c|}{Micro} & \multicolumn{3}{c}{Macro} \\
     & Precision & Recall & F1 & Precision & Recall & F1 \\ \midrule
    DMCNN & \textbf{85.88} $\pm$ 0.70 & 79.70 $\pm$ 0.59 & 82.67 $\pm$ 0.08 & 80.55 $\pm$0.49 & 73.31 $\pm$ 3.88 & 75.03 $\pm$ 0.40 \\
    BiLSTM& 83.09 $\pm$ 0.89 & 85.16 $\pm$ 0.95 & 84.11 $\pm$ 0.24 & 78.70 $\pm$ 0.92 & 76.67 $\pm$ 2.23 & 76.65 $\pm$ 1.42 \\
    BiLSTM+CRF & 84.74 $\pm$ 0.55 & 83.33 $\pm$ 0.49 & 84.03 $\pm$ 0.05 & 78.56 $\pm$ 1.31 & 72.60 $\pm$ 1.11 & 74.49 $\pm$ 0.77 \\
    BERT  & 84.19 $\pm$ 0.39 & 84.31 $\pm$ 0.34 & 84.25 $\pm$ 0.18 & 79.61 $\pm$ 0.91 & 76.76 $\pm$ 1.79 & 77.33 $\pm$ 1.30 \\
    BERT+CRF & 83.82 $\pm$ 0.48 & 84.56 $\pm$ 0.52 & 84.19 $\pm$ 0.09 & 79.77 $\pm$ 1.10 & 77.65 $\pm$ 2.20 & 77.84 $\pm$ 1.58 \\
    DMBERT & 84.77 $\pm$ 0.91 & \textbf{86.22} $\pm$ 0.77 & \textbf{85.48} $\pm$ 0.18 & \textbf{81.57} $\pm$ 1.04 & \textbf{80.90} $\pm$ 1.38 & \textbf{80.34} $\pm$ 0.74 \\
    \bottomrule
    \end{tabular}
    \caption{The test performances of ED baselines on \dataset{}. Refer to Appendix \ref{app:imple_base} for validation performances.}
    \label{tab:main}
\end{table*}

\smallskip  
\noindent
\textbf{Number of Instances}. \dataset{} is a large-scale dataset, where $89.6\%$ event types contains more than $100$ event mentions, and $43.4\%$ event types contains more than $1,000$ event mentions. Therefore, \dataset{} can provide sufficient training signals and reliable evaluation results for LED.

\section{Experiments}
\subsection{Benchmark Settings}
We randomly split the dataset into training set, validation set, and test set according to the ratio, $0.65:0.15:0.2$. Following \citet{wang2020maven}, we provide official negative samples for a fair comparison between different methods. As stated in Section~\ref{sec:cand-select}, we first employ Chinese word segmentation and POS-tagging to the documents, and then select the content words (verbs and nouns) or words in the human-collected semantic vocabulary as the trigger candidates. The detailed statistics of the data splits are listed in Table~\ref{tab:train-valid-test}. As the dataset is unbalanced, we adopt both the micro-averaged and macro-averaged precision, recall, and F1 score as the evaluation metrics for the experiments.

\subsection{Baseline Models}
Event detection has been explored for decades. In this section, we evaluate several competitive baseline models, which are widely used in the general domain event detection task, on \dataset{}, including 
(1) Token classification. We first encode the given sentences with deep neural networks, including \textbf{BiLSTM}~\cite{hochreiter1997long} and \textbf{BERT}~\cite{devlin2019bert}, and then use the hidden representations of the candidate triggers to classify their corresponding event types.
(2) Dynamic max-pooling. These models adopt convolutional neural network (\textbf{DMCNN}, \citet{chen2015event}) or pre-trained language model, BERT (\textbf{DMBERT}, \citet{wang2019adversarial}), to extract the sequence features, and employ dynamic pooling layers to obtain trigger-specific representation for each candidate.
(3) Sequence labeling. Different from previous models, we adopt sequence labelling models (\textbf{BiLSTM+CRF}, \textbf{BERT+CRF}) to capture the correlations between different events. The implementation details can be found in Appendix \ref{app:imple_base}. We run each experiment 5 times, and the averages and standard deviations of the results are reported. 

\subsection{Overall Performance Comparison}
The baseline results are shown in Table~\ref{tab:main}. And we can observe that 
(1) DMBERT can outperform other baselines significantly, with the micro-F1 score of $85.48\%$, which is still not satisfactory for real-world applications.
(2) The standard deviations on the micro-metrics are relatively small, indicating that \dataset{} contains sufficient data in the test set and can provide stable evaluation results.
(3) From the comparison between BiLSTM-based and BERT-based models, we find that BERT-based models cannot achieve significant improvement on \dataset{}. It suggests that designing event-oriented pre-trained models is necessary for LED, which we leave for future work.
(4) CRF-based models perform slightly worse than their corresponding token classification models. We attempt to employ CRF to capture the dependencies of multiple events as suggested by~\citet{wang2020maven}, while the result is inconsistent with the expectation. Therefore, it still needs exploration to model the correlations between multiple events in a single sentence.

Notably, as the legal documents are well-written and the used language is more standardized than the general domain, the event detection models can achieve better performance on \dataset{} than on the general domain dataset (DMBERT can only achieve $67.1\%$ micro-F1 score on MAVEN~\cite{wang2020maven}, while $85.5\%$ on \dataset{}.) Therefore, we can apply existing LED models to promote the downstream tasks (Section~\ref{sec:application}). However, the performance is still unsatisfactory and needs future research (Section~\ref{sec:error}).

\subsection{Error Analysis}
\label{sec:error}
To analyze the defect of existing models and point out the future directions for LED, we conduct error analysis on the prediction errors of the model with the best performance. We categorize the prediction errors into several types and find some challenges which require future efforts.

\begin{figure}[t]
    \centering
    \includegraphics[width=0.92\columnwidth]{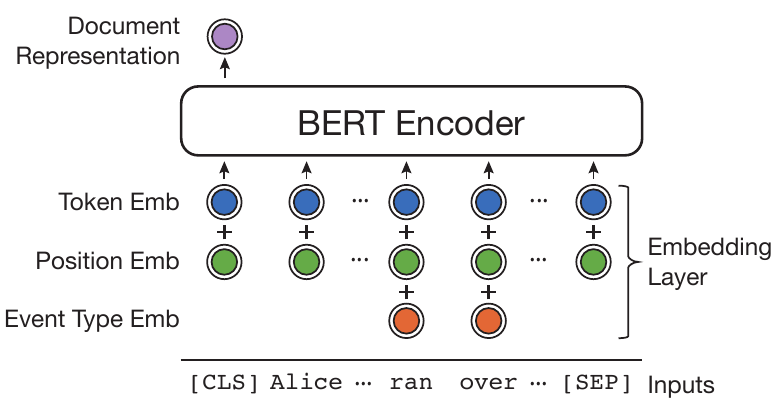}
    \caption{The framework for downstream tasks.}
    \label{fig:downstream-encoder}
\end{figure}

(1) \textbf{Long-tail Problem}. Though \dataset{} contains hundreds of thousands of event mentions, there are some event types with limited instances inevitably. We compute the performance on low-frequency event types, where the micro-F1 score is $65.97\%$ for event types with less than $50$ instances and $72.24\%$ for event types with less than $100$ instances. There is still a large gap between the performance of the low-frequency types and the overall average performance. More discussion can be found in Appendix.

(2) \textbf{Context-aware Prediction}. Many triggers require the model to integrate the information of the complex context from argument entities or other sentences to predict corresponding event types. For instance, in the sentence \textit{Bob rushed to call \textbf{ENT} to inform the situation}, if \textit{\textbf{ENT}} is \textit{the police} or \textit{110}, the event type for trigger \textit{call} is \texttt{Reporting\_to\_police}, while if \textit{\textbf{ENT}} is other people, the event type is \texttt{Reporting}. Sometimes, the essential information comes from other sentences, which require the model to capture cross sentence dependency. We randomly sample $100$ cases and ask another annotator to count the number of errors that need context-aware prediction. From the statistics, $36.98\%$ errors are caused by incorrectly capturing contextual information, which still needs further effort.

(3) \textbf{Identification Mistakes}. Similar to \citet{wang2020maven}, the most common mistake is confusing the negative samples and positive samples, i.e., the false positive and false negative. The results show that $48.99\%$ and $34.41\%$ errors are false positive and false negative, respectively. Therefore, how to identify the event semantic is a challenge.

\subsection{Applications of Legal Event Detection}
\label{sec:application}
Furthermore, in order to provide a perspective of how to use LEVEN for other Legal AI tasks and to verify the effectiveness of LED for legal documents analysis, we utilize legal events as side information in two typical downstream tasks in legal artificial intelligence, including Legal Judgment Prediction (LJP) and Similar Case Retrieval (SCR). 

In the following sections, we will first introduce the encoder architecture and applications of LED in legal judgment prediction and similar case retrieval. We compare the performance of the original BERT model and BERT model with event features to show the effectiveness of LED. Notably, the event features can either be used independently or fed into other models to further promote the performance.
The details about model implementation and dataset statistics can be found in Appendix \ref{app:imple_downs}.

\begin{table}[t]
    \centering
    \small
    \resizebox{\columnwidth}{!}{
    \begin{tabular}{l|ccc|ccc|c}
    \toprule
     \multirow{2}{*}{Model}& \multicolumn{3}{c}{Charge} & \multicolumn{3}{c}{Law} & Term \\
     & P & R & F1 & P & R & F1 & Dis $\downarrow$ \\
    \midrule  \multicolumn{8}{c}{\textit{50-shot}} \\ \midrule
    BERT          & 76.6 & \textbf{77.0} & 76.8 & 73.6 & \textbf{76.8} & 75.2 & 2.398  \\
    \quad + event & \textbf{79.2} & 76.2 & \textbf{77.7} & \textbf{75.4} & 75.6 & \textbf{75.5} & \textbf{2.364} \\ 
    \midrule  \multicolumn{8}{c}{\textit{full}} \\ \midrule
    BERT          & 88.2 & 89.4 & 88.8 & 83.7 & 86.8 & 85.2 & 1.895 \\
    \quad + event & \textbf{88.2} & \textbf{89.7} & \textbf{88.9} & \textbf{83.8} & \textbf{87.7} & \textbf{85.7} & \textbf{1.878} \\ 
    \bottomrule
    \end{tabular}}
    \caption{The results for legal judgment prediction. Here P, R, and F1 indicate precision, recall, and F1 scores, respectively.}
    \label{tab:LJP}
\end{table}

\begin{table*}[t]
    \centering
    \small
    
    \begin{tabular}{l|cccccc}
    \toprule
    Model         & MAP & NDCG@10 & NDCG@20 & NDCG@30 & P@5 & P@10\\ \midrule
    BM25          & 48.40 & 73.10 & 79.70 & 88.80 & 40.60 & 38.10 \\
    TFIDF         & 45.70 & 79.50 & 83.20 & 84.80 & 30.40 & 26.10 \\
    LMIR          & 49.50 & 76.90 & 81.80 & 90.00 & 43.60 & 40.60 \\
    Bag-of-Event  & 50.94 & 78.37 & 83.66 & 90.32 & 44.11 & 42.62 \\ 
    Bag-of-Event$_{w}$ & \textbf{51.02} & \textbf{79.90} & \textbf{84.42} & \textbf{90.97} & \textbf{45.23} & \textbf{43.36} \\ \midrule
    BERT          & 51.92 & 79.23 & 84.12 & 91.28 & 44.49 & 40.10  \\
    \quad + event & \textbf{51.99} & \textbf{80.10} & \textbf{84.92} & \textbf{91.73} & \textbf{44.63} & \textbf{41.22}  \\
    \bottomrule
    \end{tabular}
    
    \caption{The experiment results under both unsupervised and supervised settings for similar case retrieval.}
    \label{tab:case_retrieval}
\end{table*}

\subsubsection*{Encoder Architecture}
As pre-trained language models have achieved promising results in many legal tasks~\cite{chalkidis2020legal,xiao2021lawformer}, we adopt the BERT as our basic encoder. To verify the effectiveness of event detection in LegalAI, we only make minor changes in the embedding layer to integrate the event information. Figure~\ref{fig:downstream-encoder} illustrates the encoder architecture. Given the input document, to highlight the event information, we first employ the \textbf{BERT+CRF} model to detect the trigger words and their event types from the case documents\footnote{Notably, due to the high computational complexity of DMBERT, we use BERT+CRF here, which also achieve comparable results with DMBERT.}.
And then we utilize the event information in the BERT model by adding the event type embedding in the input embedding layer. The event type embedding is randomly initialized and updated during the training process. Specifically, for non-trigger tokens, we feed the sum of the token embeddings and position embeddings into the encoder. For trigger tokens, we also define an event type embedding for each event type and add the corresponding event type embeddings to the inputs.

\subsubsection*{Legal Judgment Prediction}
Legal judgment prediction (LJP) aims to predict the judgment results, including related laws, charges, and prison terms, based on the textual fact description, and LJP is an essential task for LegalAI~\cite{zhong2018legal,chalkidis2019neural,yang2019legal}. LJP requires the model to capture the key event information and mine the causal relationship between behaviors and consequences. 

Therefore, in this section, we attempt to investigate the effect of LED for judgment prediction.
We adopt the CAIL2018~\cite{xiao2018cail2018} as the evaluation benchmark, which is the largest LJP dataset. Following \citet{zhong2020does}, we formalize LJP as a multi-task learning problem. Specifically, we formalize law article prediction and charge prediction as multi-label classification tasks, and adopt binary cross-entropy function as the loss. We formalize prison term prediction as a regression task and adopt the log distance function as the loss. As for the output layer, we feed the document representation into three linear layers for the prediction of three tasks, respectively. In addition to training the model with the full dataset, we also explore the effectiveness of legal events under a low-resource setting. We only sample $50$ cases for each charge and law article to train the model.

The results are shown in Table~\ref{tab:LJP}. We can observe that LED can promote the performance of LJP, especially under low-resource settings, which proves the effectiveness of LED. Besides, LED can only achieve slight improvement on charge prediction and law prediction under with full training dataset, while can achieve consistent improvement on prison term prediction. That is because prison term prediction is more complex and requires the model to capture both the criminal behaviors and severity-level of the consequences. Legal events can provide fine-grained information for predicting prison terms, and thus promote the performance under both low-resource and full dataset settings.

\subsubsection*{Similar Case Retrieval}
Similar case retrieval (SCR) aims to retrieve supporting cases given a query case, which is a widely-applied task with high practical value~\cite{kano2018coliee,shao2021investigating}. SCR requires the leverage of  fine-grained information from multiple perspectives, including element-level, event-level, and law-level. In this paper, we adopt LeCaRD~\cite{ma2021lecard} as the evaluation benchmark, which contains $107$ queries and $43,000$ candidates. We adopt 5-fold cross-validation for evaluation, and employ the top-k Normalized Discounted Cumulative Gain (NDCG@k), Precision (P@k), and Mean Average Precision (MAP) as evaluation metrics. 

We verify the effectiveness of utilizing event features for similar case retrieval task under both unsupervised and supervised settings. In the unsupervised setting, we adopt ``Bag-of-Event'', i.e., the frequency of each event, as the representation of each legal document, and use cosine similarity to compute the similarity scores between two different legal documents. Further, considering the fact that the events occurring in the legal cases are not equally important, we compute the inverse document frequencies for different event types in the TF-IDF fashion, which are used as the weights of different event types. We denote the weighted representation as Bag-of-Event$_w$.
In the supervised setting, we train the BERT model in a sentence-pair classification paradigm. We concatenate the query case and candidate case as the input sequence, and require the model to classify whether the two cases are relevant or not.

The results are shown in Table~\ref{tab:case_retrieval}. From the results, we can observe that both Bag-of-Event and Bag-of-Event$_w$ are powerful representation methods for similar case retrieval and can achieve superior performance than other unsupervised models. Besides, the event information can facilitate the performance of BERT model, which further proves that event information is crucial for case retrieval.

\section{Conclusion and Future Work}
In this paper, we construct the largest legal event detection dataset, \dataset{}, which contains a comprehensive legal event schema and hundreds of thousands of event mentions. We evaluate several competitive baseline models and conduct error analysis for these models on \dataset{}. The experimental results prove that it still needs future efforts to promote the development of LED. Furthermore, we employ LED for downstream legal document analysis, including legal judgment prediction and similar case retrieval. It indicates that LED can provide fine-grained information and serve as a fundamental process for legal artificial intelligence. In the future, we will explore to conduct more analysis on large-scale legal documents based on the event information, and annotate \dataset{} with event relations and event arguments.

\section*{Ethical Considerations}
LEVEN focuses on detecting events from the fact and does not involve any value judgment. LED aims to transform the unstructured legal text into structured event information, which is helpful to further processing. Therefore, our work can help reduce the workload for legal professionals and improve their work efficiency. Considering the fact that, like any other legal AI application, LED models would inevitably make mistakes and have negative influences, we argue that LED can only serve as an auxiliary tool for legal work and the final decision on a specific legal issue has to be ensured by legal professionals. In such case, we could exploit the advantage of legal AI and avoid the potential risk. 

The corpus we use is released by the Chinese government and has been anonymized wherever necessary. Therefore, our dataset does not involve any personal privacy. In terms of human annotation, we first annotate a few examples on our own to approximate the workload and then determine the wages for annotators according to local standards.

\section*{Acknowledgement}
We are grateful to Meng Zhang and Danyang Guo from School of Law, Tsinghua University for their professional legal support. This work is supported by the National Key R\&D Program of China (No. 2018YFC0831900, No. 2020AAA0106502), Institute for Guo Qiang at Tsinghua University, Beijing Academy of Artificial Intelligence (BAAI), and International Innovation Center of Tsinghua University, Shanghai, China.

Feng Yao, Chaojun Xiao, and Xiaozhi Wang designed the annotation schema. 
Feng Yao, Chaojun Xiao, and Xiaozhi Wang designed and conducted the experiments.
Chaojun Xiao, Xiaozhi Wang, Feng Yao, and Zhiyuan Liu wrote the paper. Lei Hou, Cunchao Tu, Juanzi Li, Yun Liu, Weixing Shen, and Maosong Sun provided valuable advice to the research.

\bibliography{anthology,custom}
\bibliographystyle{acl_natbib}

\appendix

\section{Implementation Details}
\label{app:imple_det}
In this section, we introduce the hyper-parameters of the baseline models, models for legal judgment prediction, and models for similar case retrieval. 

\subsection{Baseline Models}
\label{app:imple_base}
For all baseline models, we run the model $5$ times to get stable results, and the average performance is reported. For each model, we choose the checkpoint with the best performance on the validation set to evaluate on the test set. We train these models on GeForce RTX 2080Ti GPUs, and use Adam to optimize the models. The validation performances are shown in Table \ref{tab:valid-result}.

\begin{table}[h]
    \centering
    \small
    \resizebox{\columnwidth}{!}{
    \begin{tabular}{l|ccc|ccc}
    \toprule
    \multirow{2}{*}{Model} & \multicolumn{3}{c|}{Micro} & \multicolumn{3}{c}{Macro} \\
     & P & R & F1 & P & R & F1 \\ \midrule
    DMCNN        & 86.15 & 79.27 & 82.57 & 79.42 & 69.77 & 73.00 \\
    BiLSTM       & 83.01 & 84.30 & 83.65 & 78.45 & 73.39 & 74.27 \\
    BiLSTM+CRF   & 84.63 & 83.10 & 83.86 & 80.99 & 73.39 & 75.73 \\
    BERT         & 84.35 & 83.80 & 84.07 & 80.21 & 76.08 & 77.38 \\
    BERT+CRF     & 83.72 & 84.13 & 83.93 & 78.38 & 75.39 & 76.01 \\
    DMBERT       & 83.40 & 86.76 & 85.05 & 79.18 & 79.28 & 78.42 \\
    \bottomrule
    \end{tabular}
    }
    \caption{Validation Performances of ED baselines.}
    \label{tab:valid-result}
\end{table}

For \textbf{DMCNN}, the hyper-parameters are the same as \citet{chen2015event}, excluding the unmentioned dimension of word embedding and learning rate. We use JIEBA toolkit\footnote{\url{https://github.com/fxsjy/jieba}} to perform the Chinese word segmentation, and use the pre-trained word vectors released in \citet{wordemb}. Specifically, the word embedding is the one trained by the Wikipedia-zh corpus with word, character, and N-gram context. The training parameters are shown in Table \ref{tab:dmcnn-hyper}.

\begin{table}[h]
    \centering
    \small
    \resizebox{5.5cm}{!}{
    \begin{tabular}{l|c}
    \toprule
        Batch Size             & 170   \\
        Dropout Rate           & 0.5   \\
        Learning Rate          & $1\times10^{-3}$ \\
        Kernel Size            & 3     \\
        Hidden Size            & 200   \\
        Dimension of PF        & 5     \\
        Dimension of Word Embedding   & 300   \\ 
        \bottomrule
    \end{tabular}
    }
    \caption{Hyper-parameters for DMCNN.}
    \label{tab:dmcnn-hyper}
\end{table}

For \textbf{BiLSTM-based models}, we also use JIEBA to perform word segmentation, and adopt the same pre-trained Chinese word vectors used in DMCNN. The detailed training hyper-parameters are shown in Table
\ref{tab:bilstm-hyper}.

\begin{table}[h]
    \centering
    \small
    \resizebox{5.5cm}{!}{
    \begin{tabular}{l|c}
    \toprule
        Batch Size       & 200 \\
        Dropout Rate     & 0.5 \\
        Learning Rate    & $1\times10^{-3}$ \\
        Kernel Size      & 3  \\
        Hidden Size      & 256 \\
        Dimension of Word Embedding & 300  \\ 
        \bottomrule
    \end{tabular}
    }
    \caption{Hyper-parameters for BiLSTM-based models.}
    \label{tab:bilstm-hyper}
\end{table}
For \textbf{BERT-based models}, we adopt BERT-base as the basic encoder with the bert-base-chinese checkpoint\footnote{\url{https://huggingface.co/bert-base-chinese}}. For BERT, BERT-CRF and DMBERT, the training hyper-parameters are shared and the training hardwares are 4$\times$ RTX 2080TI. The detailed hyper-parameters are shown in Table \ref{tab:bert-hyper}.

\begin{table}[h]
    \centering
    \small
    \resizebox{5.5cm}{!}{
    \begin{tabular}{l|c}
    \toprule
        Batch Size       & 64 \\
        Dropout Rate     & 0.5 \\
        Adam $\epsilon$  & $1\times10^{-8}$ \\
        Learning Rate    & $5\times10^{-5}$ \\
        Validation Steps During Training & 500  \\
        \bottomrule
    \end{tabular}
    }
    \caption{Hyper-parameters for BERT-based models.}
    \label{tab:bert-hyper}
\end{table}

For \textbf{CRF-based models}, we use BIO tagging schema for training and evaluation.

\subsection{Downstream Applications}
\label{app:imple_downs}
For all downstream application experiments, we first adopt the \textbf{BERT+CRF} model to detect the trigger words in the original downstream datasets. We add an extra \textbf{Event Type Embedding Layer} to incorporate the event information, where the embedding matrix with the shape of $109\times768$ is randomly initialized and updated during training.

For \textbf{Legal Judgment Prediction (LJP)} task, we adopt the dataset released in the first stage of CAIL2018\footnote{\url{https://github.com/china-ai-law-challenge/CAIL2018}} as the benchmark. We evaluate the LJP task with the detected events under both full data and low-resource settings. The data size is shown in Table \ref{tab:data-ljp}. The training data for the low-resource setting is obtained by randomly selecting 50 samples for each label and the corresponding data we used is also released in the github repository. The experiments are based on the source code released in \citet{zhong2020does}.

\begin{table}[h]
    \centering
    \small
    \resizebox{\columnwidth}{!}{
    \begin{tabular}{c|rcc}
    \toprule
        Setting       &   Training     & Validation      & Test \\ \midrule
        Full-data     &   154,592     & \multirow{2}{*}{17,131} & \multirow{2}{*}{32,508}\\
        Low-resource  &    12,702     \\
    \bottomrule
    \end{tabular}
    }
    \caption{Statistics of Data for LJP Experiments.}
    \label{tab:data-ljp}
\end{table}


For \textbf{Similar Case Retrieval (SCR)}, we use LeCaRD\footnote{\url{https://github.com/myx666/LeCaRD/tree/main/data}} as benchmark and implement the models based on the code released in \citet{xiao2021lawformer}. As the documents in LeCaRD are usually longer than $512$, we truncate the text to feed into the encoder. Specifically, the maximum lengths we use for the query and candidates are 100 and 409, respectively. We also conduct experiments in the unsupervised setting, where we use a 108-dimension vector as the representation of a document and each entry of the vector is the number of the detected events.


\section{Performance on Different Top-level Types}

\begin{table}[h]
    \centering
    \small
    \resizebox{\columnwidth}{!}{
    \begin{tabular}{c|ccc}
    \toprule
        \makecell{Top-level Event Type} & precision & recall & F1 \\ \midrule
        \texttt{General\_behaviors}      & 83.71 & 85.67 & 84.86 \\
        \texttt{Prohibited\_acts}     & 83.01 & 82.93 & 82.97 \\
        \texttt{Judicature\_related}    & 94.17 & 91.89 & 93.01  \\
        \texttt{Consequences}            & 84.54 & 82.92 & 83.73  \\
        \texttt{Accident}               & 86.04 & 84.40 & 85.21  \\ 
        \texttt{Natural\_disaster}      & 77.78 & 63.64 & 70.00  \\ 
        \bottomrule
    \end{tabular}
    }
    \caption{Performance of DMBERT on different top-level event types.}
    \label{tab:top-level-performance}
    \vspace{-0.5em}
\end{table}

In this section, we further analyze the performance of DMBERT on different top-level types to explore the fine-grained performance on LEVEN. From the results, we can observe that as the \texttt{Judicature\_related} events are usually described in legal terminologies, thus the models can easily identify trigger words correctly by memorizing a set of specific words. As the \texttt{Natural\_disaster} only contains tens of event instances, the model cannot be well-trained for these event types.

\section{Performance on Long-tail Event Types}

In this section, we further analyze the performance on long-tail event types in detail. From Table~\ref{tab:long-tail-eval}, we can observe that overall, the model performance decreases as the number of training instances decreases. But there are some exceptions. Some event types only contain limited instances while the model can achieve high F1-scores on these types. 
For instances, \texttt{Suicide} only contains $55$ event mentions, but the model can achieve $95.65 \%$ F1-score due to its non-diverse expression. Though some long-tail event types can be predicted accurately, there are 9 long-tail event types that can only reach F1-scores lower than $0.6$. Therefore, we argue that detecting the event types accurately with limited instances needs future efforts.

\begin{table}[h]
    \centering
    \small
    \resizebox{\linewidth}{!}{
    \begin{tabular}{c|ccccc|c}
    \toprule
         F1-score           & [0,0.4) & [0.4,0.6) & [0.6,0.8) & [0.8,0.9) & [0.9,1.0] & sum\\ \midrule
            \#low-freq.     & 5       &    4      &    4      &  4        & 4         & 21  \\
            \#mid-freq.     & 0       &    0      &    9      &  13       & 6         & 28\\
            \#high-freq.    & 0       &    0      &    14     &  23       & 22        & 59 \\ \bottomrule
    \end{tabular}
    }
    \caption{Distribution of event types by their performance on the test set. Here, low-freq and high-freq represent the number of event types that have less than 150 event mentions and more than 500 event mentions. And mid-freq denotes the number of event types containing between 150 and 500 event mentions.}
    \label{tab:long-tail-eval}
\end{table}
\smallskip
\noindent




\section{Data Distribution}
To help the following researchers to better understand the features and details of \dataset{}, we present more data analysis in this section regarding multiple events in one sentence and the sentence length distribution.

\smallskip
\noindent
\textbf{Number of Events in One Sentence}. Legal cases usually involve complicated facts, and it is common in \dataset{} that there are multiple events mentioned in one sentence. Table~\ref{tab:multi-eve} shows the percentage of sentences containing different numbers of events. 

\begin{table}[h]
    \centering
    \small
    \resizebox{\linewidth}{!}{
    \begin{tabular}{c|ccccc}
    \toprule
         \#Event/Sent.  & 0 & 1 & [2,5) & [5,10) & [10,100) \\ \midrule
         Percentage (\%)& 12.8 & 26.7 & 47.9 & 11.6 & 1.0 \\ \bottomrule
    \end{tabular}
    }
    \caption{The percentage of sentences containing different numbers of sentences.}
    \label{tab:multi-eve}
\end{table}
\smallskip
\noindent

\begin{figure}[ht]
    \centering
    \includegraphics[width=\columnwidth]{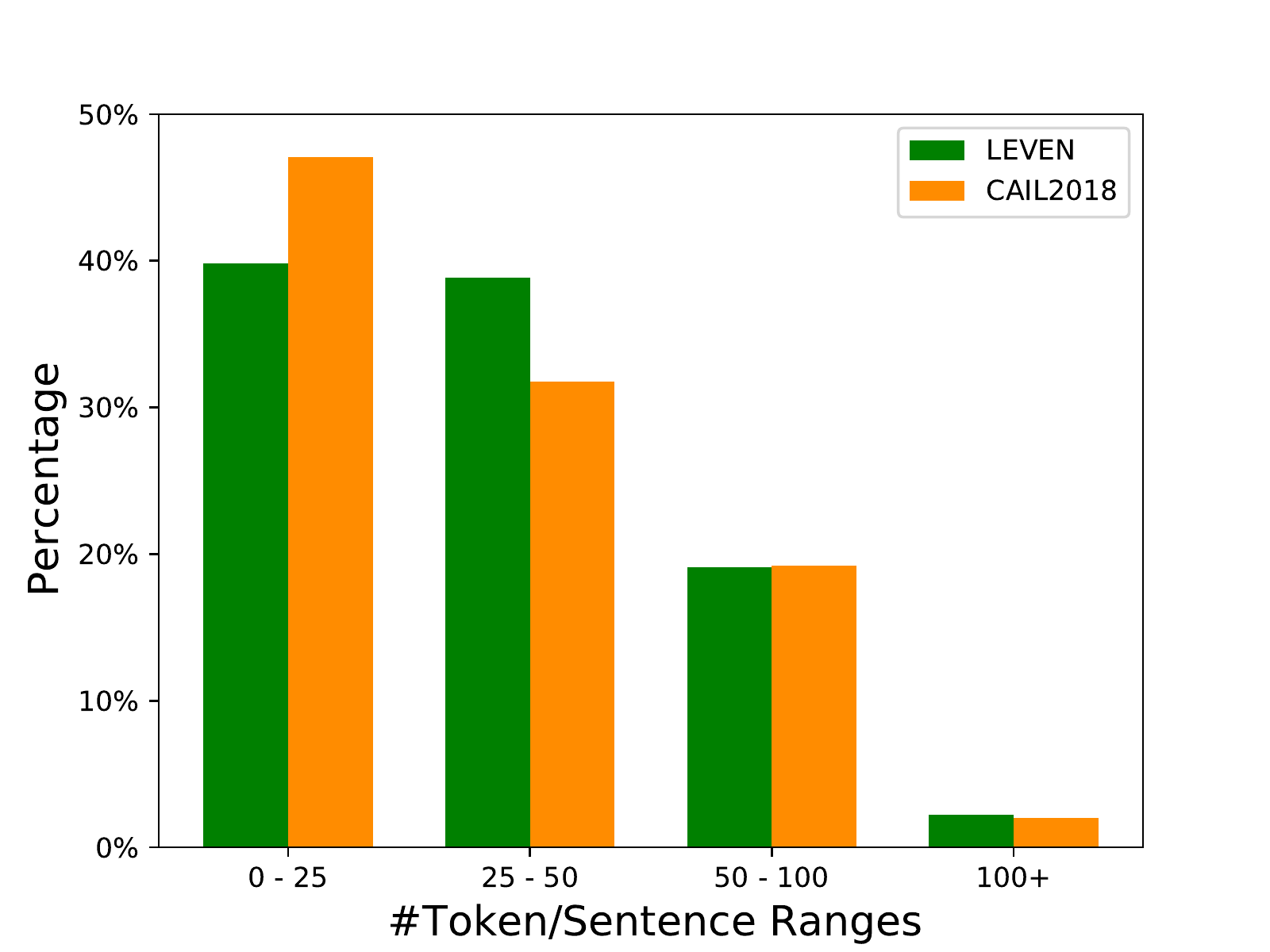}
    \caption{Sentence length distributions of LEVEN and CAIL2018.}
    \label{fig:sent_len}
\end{figure}

\begin{figure}[ht]
    \centering
    \includegraphics[width=\columnwidth]{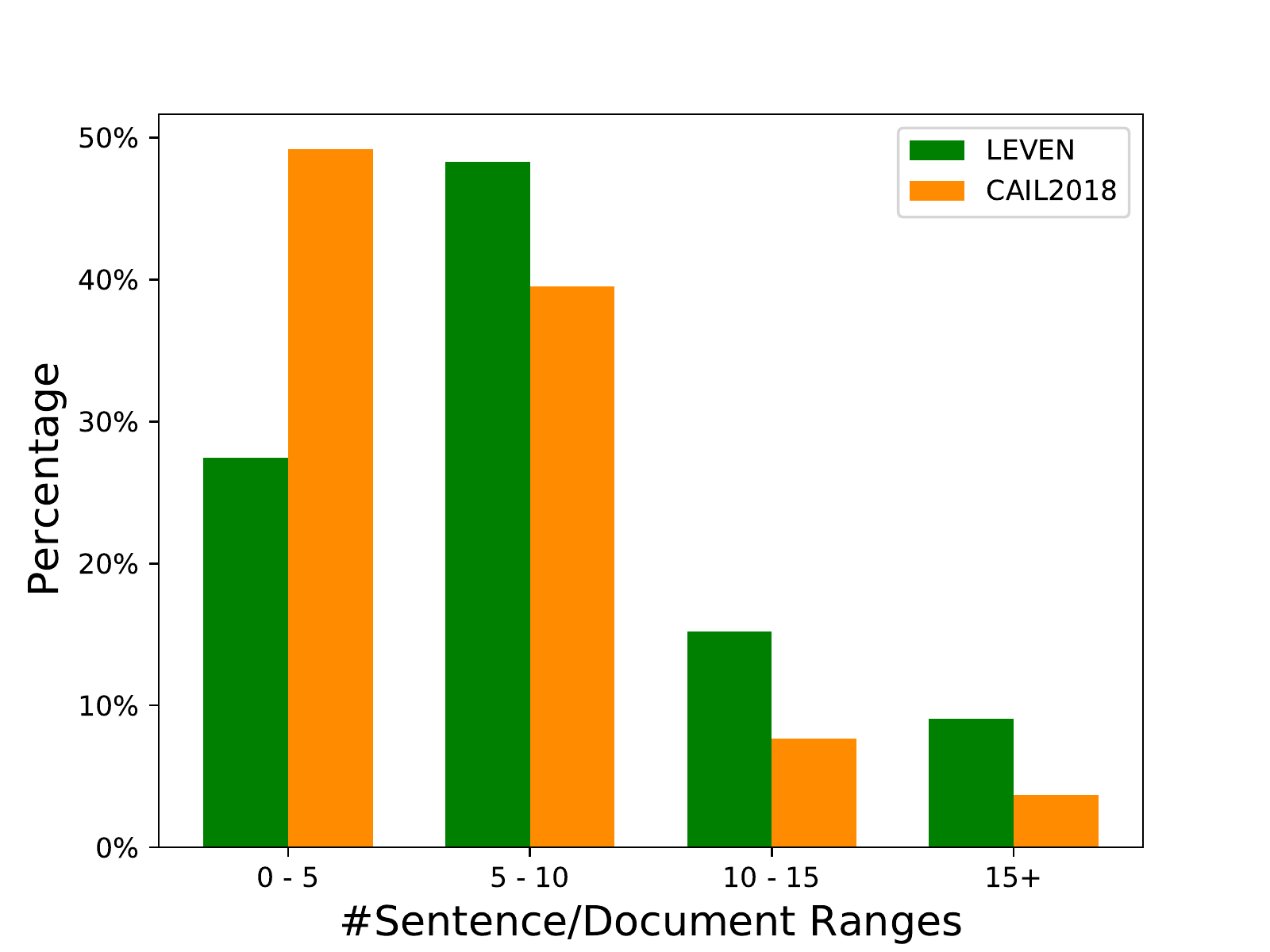}
    \caption{Sentence number distributions of LEVEN and CAIL2018.}
    \label{fig:sent_num}
\end{figure}

\smallskip  
\noindent
\textbf{Length\&Number of Sentences}. \dataset{} is constructed based on real-world corpus, which makes it a perfect resource for developing practical applications. Figure \ref{fig:sent_len} and \ref{fig:sent_num} exhibit a comparison between the sentence length and number distributions of \dataset{} and CAIL2018\citep{xiao2018cail2018}, which is the largest legal judgment prediction dataset with over 1.7 million criminal judgment documents and can serve as a good real-world reference, indicating that \dataset{} is consistent with the reality.

\section{Event Type Schema and Description}
\label{app:event_schema}
\begin{figure*}[t]
    \centering
    \includegraphics[height=0.98\textheight]{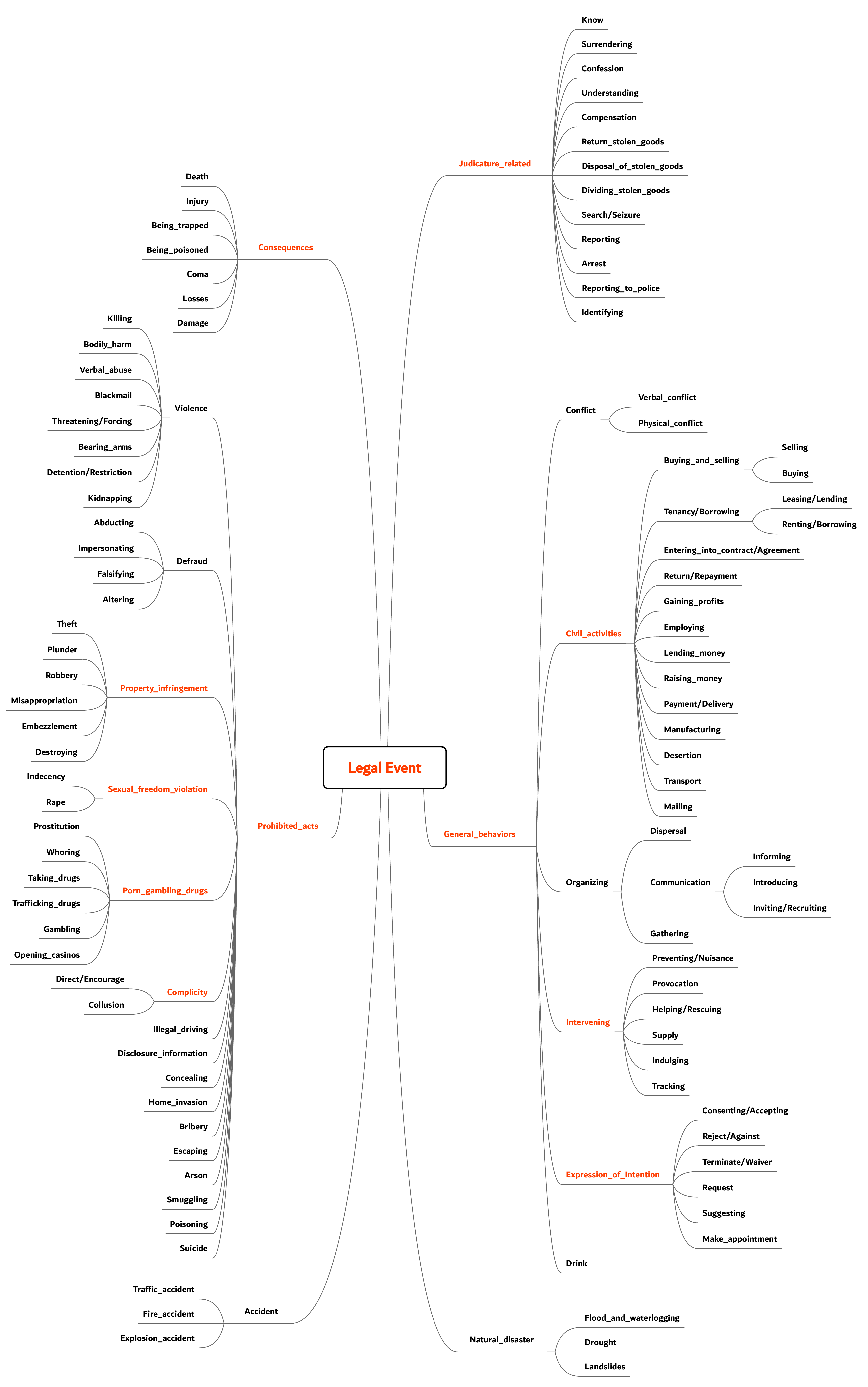}
    \caption{The detailed event schema of our proposed \dataset{}.}
    \label{fig:schema}
\end{figure*}

To promote future research, we provide the hierarchical event schema in Figure~\ref{fig:schema}, and the list of event types, including the event names and the corresponded descriptions, in Table~\ref{tab:type-des1},~\ref{tab:type-des2},~\ref{tab:type-des3}, and~\ref{tab:type-des4}.

\section{Annotation Guidelines}
\label{app:annote_guide}
The annotation guidelines can be obtained from \url{https://github.com/thunlp/LEVEN}.

\newcommand{\RNum}[1]{\uppercase\expandafter{\romannumeral #1\relax}}
\begin{table*}[ht]
    \centering
    \small
    \begin{tabularx}{\textwidth}{l|X}
    \toprule
    Event Type Name & Descriptions \\ \midrule
    \multicolumn{2}{c}{Judicature Related Events} \\ \midrule
    Judicature\_related  & JUDICATURE\_RELATED events mainly refer to the activities of judicial organs or some legal penalty circumstances. \\
    Know     & A KNOW event means the doer ought to know the fact or understand the fact clearly. \\
    Surrendering & A SURRENDERING event refers to the doer voluntarily surrendering after committing a crime. \\
    Confession & A CONFESSION event refers to the suspect or defendant telling the facts to the police. \\
    Understanding & AN UNDERSTANDING event refers to the forgiveness from the victim or victim's families to the criminal. \\
    Compensation & A COMPENSATION event refers to the act of compensating the victim for his loss, damage, or injury. \\
    Return\_stolen\_goods & A TERURN\_STOLEN\_GOODS event refers to the act of returning the stolen money or stolen goods to the victim or government. \\
    Disposal\_of\_stolen\_goods & A DISPOSAL\_OF\_STOLEN\_GOODS event refers to the act of destroying stolen goods, selling stolen goods, or squandering stolen money, that is, the stolen goods/money have been disposed of. \\
    Dividing\_stolen\_goods & A DICIDING\_STOLEN\_GOODS event refers to the act of sharing stolen goods or money. \\
    Search/Seizure & A SEARCH/SEIZURE event mainly refers to the search and inspection of the suspect's body, articles, residence, or other space by the reconnaissance personnel, or the seizure of contraband, including the seizure of real estate. However, illegal search or seizure by non-reconnaissance personnel can also mark this event. \\
    Reporting & A REPORTING event refers to the act of reporting bad people or bad things to relevant units. \\
    Arrest & AN ARREST event refers to the act of detaining or arresting suspects. \\
    Reporting\_to\_police & A REPORTING\_TO\_POLICE event refers to the act of calling the police to ask for help or reporting a case to the police. \\
    Identifying & AN IDENTIFYING event refers to a kind of behavior in which the investigation organ appoints or hires people with expertise to make a scientific judgment and draw professional conclusions on the specialized problems in criminal cases in order to solve the specialized problems in criminal cases. \\
    
    \midrule
    \multicolumn{2}{c}{Accident Events} \\ \midrule
    Accident & AN ACCIDENT event refers to accidental loss or disaster. \\
    Traffic\_accident & A TRAFFIC\_ACCIDENT event occurs when a traffic accident happens, which usually causes personal injury, death or property loss. \\
    Fire\_accident & A FIRE\_ACCIDENT event refers to the disaster caused by uncontrolled combustion. \\
    Explosion\_accident & AN EXPLOSION\_ACCIDENT event refers to the disaster caused by a sudden release of a large amount of energy, which leads to property losses and personal casualties. \\
    \midrule
    \multicolumn{2}{c}{Natural Disaster Events} \\ \midrule
    Natural\_disaster & NATURAL\_DISASTER events refer to Natural phenomena or man-made influences that endanger human survival or damage the human living environment. \\
    Flood\_and\_waterlogging & A FLOOD\_AND\_WATERLOGGING event occurs where a large amount of water covers an area that is usually dry. \\
    Drought & A DROUGHT event occurs when there is little or no rain during a long period of time. \\
    Landslides & A LANDSLIDES event refers to a geographic disaster caused by a mass of earth or rock falling down the slope of a mountain. \\
    
    \midrule
    \multicolumn{2}{c}{Consequence Events} \\ \midrule
    Consequence & CONSEQUENCE events contain the fact of damage to the object caused by  harmful acts. \\
    Death & A DEATH event refers to the state of a human being dead. \\
    Injury & AN INJURY event refers to the fact of personal injury. \\
    Being\_trapped & A BEING\_TRAPPED event means the state in which people are physically in trouble and can't get out. \\
    Being\_poisoned & A BEING\_POISONED event refers to one's discomfort caused by toxic effects, emphasizing the state of one's being poisoned. \\
    Coma & A COMA event refers to the state of one's unconsciousness. \\
    Losses & A LOSSES event refers to the fact of property loss. \\
    Damage & A DAMAGE event refers to the fact that the property has been damaged. \\
    
    \midrule
    \multicolumn{2}{c}{General Behavior Events (\RNum{1})} \\ \midrule
    General\_behavior & GENERAL\_BEHAVIOR events contain common behaviors in daily life,  which usually do not violate laws. \\
    Conflict & A CONFLICT event refers to two or more parties having verbal, physical, or other conflicts, disputes, or contradictions. \\
    \bottomrule
    \end{tabularx}
    \caption{Event type list (\RNum{1}), including the event type names and the corresponded descriptions.}
    \label{tab:type-des1}
\end{table*}

\begin{table*}[ht]
    \centering
    \small
    \begin{tabularx}{\textwidth}{l|X}
    \toprule
    Event Type Name & Descriptions \\ \midrule
    \multicolumn{2}{c}{General Behavior Events (\RNum{2})} \\ \midrule
    Verbal\_conflict & A VERBAL\_CONFLICT event refers to oral conflicts happen between two or more people without physical contact. \\
    Physical\_conflict & A PHYSICAL\_CONFLICT event refers to a physical clash that happens between two or more people, including fighting. This event emphasizes the mutual behavior of both parties, pay attention to distinguish this event from a BODILY\_HARM event, which emphasizes that one  hurts another. \\
    Civil\_activities & CIVIL\_ACTIVITIES events contain typical activities in civil and commercial areas. \\
    Buying\_and\_selling & A BUYING\_AND\_SELLING event refers to the act of transacting within or between groups, including the exchange of goods and online transactions. \\
    Selling & A SELLING event refers to one's act of selling something for a profit. \\
    Buying & A BUYING event refers to one's act of buying or consuming something. \\
    Tenancy/Borrowing & A TENANCY/BORROWING event refers to the relationship between two groups/persons to lease or rent something. \\
    Leasing/Lending & A LEASING/LENDING event refers to the act of renting or lending something to others. \\
    Renting/Borrowing & A RENTING/BORROWING event refers to the act of renting or borrowing something from others. \\
    Return/Repayment & A RETURN/REPAYMENT refers to the act of returning something to its original place or owner. \\
    Gaining\_profits & A GAINING\_PROFITS event refers to one obtaining money or other benefits through a certain act or activity. \\
    Employing & AN EMPLOYING event refers to the act of giving others a job to do for payment. \\
    Lending\_money & A LENDING\_MONEY event refers to specialized institutions or people making loans to earn profits, including bank loans and individual loans. \\
    Raising\_money & A RAISING\_MONEY event refers to the act of raising money from unspecified majority people. \\
    Payment/Delivery & A PAYMENT/DELIVERY event refers to the act of giving money or other things to others. \\
    Entering\_into\_contract/agreement & AN ENTERING\_INTO\_CONTRACT/AGREEMENT event refers to the act of two or more person/groups signing contracts, including written contracts, written agreements, oral agreements, etc. \\
    Manufacturing & A MANUFACTURING event refers to producing, manufacturing, or making tangible objects, emphasizing from scratch, excluding "noise", "explosion" or other intangible objects. \\
    Desertion & A DESERTION event refers to one's act of actively abandoning or discarding something or someone. \\
    Transport & A TRANSPORT event refers to one's act of transporting someone or something from one place to another. \\
    Mailing & A MAILING event refers to delivering documents or articles through the post office or third-party postal service. \\
    Organizing & AN ORGANIZING event refers to the act of arranging scattered people or things to serve a common goal. \\
    Dispersal & A DISPERSAL event refers to the act of spreading information, data, rumors to the unspecified majority of people on the Internet or in public. \\
    Communication & A COMMUNICATION event generally refers to the connection between two or more people, such as making a phone call. \\
    Informing & AN INFORMING event refers to one's act of telling others information or reminding others of certain information, or the notified one should not have known the information. \\
    Introducing & AN INTRODUCING refers to one's behavior to make other people or groups  know each other or have a connection, excluding product instructions (because the introduction here does not mean ``intermediary", but just a kind of teaching). \\
    Inviting/Recruiting & AN INVITING/RECRUITING event refers to the acts of recruiting, inviting others to a place, or inviting others to do something or participate in an activity. \\
    Gathering & A GATHERING event refers to the act of gathering a group of people together. \\
    Intervening & AN INTERVENING event refers to one's act of intervening in an ongoing event. \\
    Preventing/Nuisance & A PREVENTING/NUISANCE event refers to one's act of preventing things from going smoothly or hindering others from doing something by words or actions. \\
    Provocation & A PROVOCATION event refers to one attempting to trigger off conflicts with others, or trigger off conflicts between other two groups. \\

    \bottomrule
    \end{tabularx}
    \caption{Event type list (\RNum{2}), including the event type names and the corresponded descriptions.}
    \label{tab:type-des2}
\end{table*}

\begin{table*}[ht]
    \centering
    \small
    \begin{tabularx}{\textwidth}{l|X}
    \toprule
    Event Type Name & Descriptions \\ \midrule
    \multicolumn{2}{c}{General Behavior Events (\RNum{3})} \\ \midrule
    Helping/Rescuing & A HELPING/RESCUING event refers to one's act of helping others to do something in the process of life, work or crime, it is limited to behavioral help, excluding providing materials, suggestions, etc. A HELPING/RESCUING event also refers to one's act of saving, rescuing, or assisting others who are injured or in trouble. \\
    Supply & A SUPPLY event refers to one providing materials, conditions, intelligence information, or other specific things to others, excluding abstract things such as ``providing help" or ``providing advice". \\
    Indulging & AN INDULGING event refers to one's act of allowing bad things to develop without any interference. \\
    Tracking & A TRACKING event refers to one's act of following others quietly without being detected. \\
    Expression\_of\_Intention & EXPRESSION\_OF\_INTENTION events contain the acts of one expressing a certain intention in a verbal way. \\
    Consenting/Accepting & A CONSENTING/ACCEPTING event refers to one agreeing with the opinions of others, accepting others' asks, or accepting the property given by others. \\
    Reject/Against & A REJECT/AGAINST event refers to one rejecting others' asks or the property given by others. \\
    Terminate/Waiver & A TERMINATE/WAIVER event refers to one stopping doing something, giving up the original persistence, or giving up a right. \\
    Request & A REQUEST event refers to one putting forward specific matters or wishes, hoping or requiring others to realize them. \\
    Suggesting & A SUGGESTING event refers to one putting forward a plan or idea to others. \\
    Make\_appointment & A MAKE\_APPOINTMENT event refers to the act of two or more people discussing and determining something. \\
    Drink & A DRINK event refers to one's act of drinking alcohol, usually accompanied by other behaviors, such as driving, etc. \\
    
    \midrule
    \multicolumn{2}{c}{Prohibited Acts Events (\RNum{1})} \\ \midrule
    Prohibited\_acts & PROHIBITED\_ACTS events contain behaviors prohibited by law, including not only typical criminal behaviors, but also behaviors that are not up to the degree of crime but prohibited by law. Therefore, events in this part are events that should be given negative evaluation, which is opposite to general behaviors. \\
    Violence & VIOLENCE events contain violent behaviors that are intended to hurt others' mental or physical health, including physical force as well as language. \\
    Killing & A KILLING event refers to one's act of killing others in order to make others die. \\
    Bodily\_harm & A BODILY\_HARM event refers to the act of harming the physical health of others, usually manifested in beating. \\
    Verbal\_abuse & A VERBAL\_ABUSE event refers to the act of insulting, attacking or hurting others through language. Pay attention to distinguishing this event from a VERBAL\_CONFLICT event, which emphasizes mutual abuse. \\
    Blackmail & A BLACKMAIL event refers to the act of demanding money from others by threatening or deceiving them. \\
    Threatening/Forcing & A THREATENING/FORCING event refers to the act of forcing others to do or not do something through violence or power, mostly referring to the use of force to make others obey. \\
    Bearing\_arms & A BEARING\_ARMS event refers to one's holding or carrying sticks, props, guns, or other instruments. \\
    Detention/restriction & A DETENTION/RESTRICTION event refers to the act of depriving or restricting the freedom of others, such as binding or detaining people in specific places. \\
    Kidnapping & A KIDNAPPING event refers to the act of taking hostages by violent means in exchange for interests, emphasizing that the object must be people. \\
    Defraud & A DEFRAUD event refers to the act of covering up the real situation with false words or actions to deceive others. \\
    Abducting & AN ABDUCTING event refers to one's act of cheating someone away by luring, cheating, or other means. \\
    Impersonating & AN IMPERSONATING event refers to the act of disguising a real thing with a false thing or one's act of pretending to be somebody in order to trick people. \\
    Falsifying & A FALSIFYING event refers to the act of making fake goods or false news. \\
    Altering & AN ALTERING event refers to the act of modifying real basis A without authorization to make it have another illusion B. \\
    Property Infringement & PROPERTY INFRINGEMENT events contain acts of infringing upon others' property rights and interests of others. \\
    Theft & A THEFT event refers to one's act of stealing others' property by secret. \\
    Plunder & A PLUNDER event refers to one's act of seizing property blatantly in front of the victims and taking them away, including seizing guns or knives, excluding competing for customers or land rights. The object of robbery must be tangible things. \\

    \bottomrule
    \end{tabularx}
    \caption{Event type list (\RNum{3}), including the event type names and the corresponded descriptions.}
    \label{tab:type-des3}
\end{table*}

\begin{table*}[ht]
    \centering
    \small
    \begin{tabularx}{\textwidth}{l|X}
    \toprule
    Event Type Name & Descriptions \\ \midrule
    \multicolumn{2}{c}{Prohibited Acts Events (\RNum{2})} \\ \midrule
    Robbery & A ROBBERY event refers to one's act of using violent means to rob others' property, such as robbery with a knife. The establishment of this event is strict. If it is impossible to judge whether it is a ROBBERY event, then PLUNDER may be marked. \\
    Misappropriation & A MISAPPROPRIATION event refers to the act of changing the original use of the property to another without authorization. \\
    Embezzlement & AN EMBEZZLEMENT event refers to one's act of taking others' property illegally, including real estate, emphasizing the state of possession. \\
    Destroying & A DESTROYING event refers to one's act of destroying property, this event has a subject, which is the main difference against A DAMAGE event. \\
    Sexual\_freedom\_violation & SEXUAL\_FREEDOM\_VIOLATION events contain acts of making others unable to freely dispose of their sexual rights by means of inducement, deception, coercion, etc. \\
    Indecency & AN INDECENCY event refers to one's act of forcibly sexually harassing others by touching private parts or other acts other than adultery. \\
    Rape & A RAPE event refers to one's act of forcing women to have sex when they do not want to. \\
    Porn\_gambling\_drugs & PORN\_GAMBLING\_DRUGS events contain illegal or criminal phenomena involving pornography, gambling, and drugs. \\
    Prostitution & A PROSTITUTION event refers to women providing paid sexual services to others. \\
    Whoring & A WHORING event refers to one purchasing sexual service with money. \\
    Taking\_drugs & A TAKING\_DRUGS event refers to one's act of taking drugs. \\
    Trafficking\_drugs & A TRAFFICKING\_DRUGS event refers to one's act of peddling drugs. \\
    Gambling & A GAMBLING event refers to one's act of gambling. \\
    Opening\_casinos & AN OPENING\_CASINOS event refers to one's act of opening casinos for multiple plays to gamble on. \\
    Complicity & COMPLICITY events occur when intentional contacts happen between two or more criminals. \\
    Direct/Encourage & A DIRECT/ENCOURAGE refer to one's act of letting others commit crimes by means of command, inspiration, or temptation. Specifically, a DIRECT event refers to the act of summoning others to commit criminal acts or other negative acts according to the instigator's intention. AN ENCOURAGE event refers to one's act of making people who do not have criminal intention have the intention of committing a crime. \\
    Collusion & A COLLUSION event refers to the act of two or more people scheming a crime plan together. \\
    Illegal\_driving & AN ILLEGAL\_DRIVING event refers to one's act of driving a car illegally. \\
    Disclosure\_information & A DISCLOSURE\_INFORMATION event refers to one's act of disclosing information that should be kept secret. \\
    Concealing & A CONCEALING event refers to one's act of hiding something from discovery. \\
    Home Invasion & A HOME INVASION event refers to one's act of invading or sneaking into other people's private space without the permission of others. This event is usually the pre-act of another criminal act (such as theft or rape). \\
    Bribery & A BRIBERY event refers to one's act of bribing others with property to seek illegitimate interests or accepting others' property to seek illegitimate interests for others. \\
    Escaping & AN ESCAPING event means one's escaping and hiding in order to avoid capture. \\
    Arson & AN ARSON event refers to one's act of setting on fire. \\
    Smuggling & A SMUGGLING event refers to the act of one's illegally transporting goods into or out of the country in violation of customs regulations. \\
    Poisoning & A POISONING event refers to one putting poison in containers or a specific environment in order to kill people, animals, or plants. \\
    Suicide & A SUICIDE event refers to the act of one's killing himself. \\
    \bottomrule
    \end{tabularx}
    \caption{Event type list (\RNum{4}), including the event type names and the corresponded descriptions.}
    \label{tab:type-des4}
\end{table*}

\end{document}